\documentclass{clv3}

\usepackage{hyperref}
\usepackage{xcolor}

\usepackage{times}
\usepackage{latexsym}
\usepackage{amsmath}
\usepackage{amsthm}
\usepackage{amsfonts}
\usepackage{amssymb}
\usepackage{graphicx}
\usepackage{booktabs} 

\usepackage{verbatim}
\usepackage{url}

\newcommand\lvec[1]{\overrightarrow{\mathbf{#1}}}
\newcommand\rvec[1]{\overleftarrow{\mathbf{#1}}}
\newcommand*\concat{\hspace{0.2em}\mathbin{\|}\hspace{0.2em}}

\renewcommand\vec[1]{\mathbf{#1}}

\newcommand\matrow[3][]{{#2}{[#3]}^{#1}}

\newcommand\RR{\mathbb{R}}
\newcommand{\nth}[1]{${#1}^\text{th}$}

\newcommand{\ere}{$\mathcal{E}\mbox{2}\mathcal{E}$RE}

\newcommand\set[1]{\mathcal{#1}}

\newcommand\ci[1]{\raisebox{.5ex}{\scriptsize $\pm$ #1}}

\usepackage{tikz,pgfplots}
\newcommand*\circled[1]{\tikz[baseline=(char.base)]{
            \node[shape=circle,draw,inner sep=1pt] (char) {#1};}}

\newcommand\LSTMR{\overrightarrow{\text{LSTM}}}
\newcommand\LSTML{\overleftarrow{\text{LSTM}}}

\newcommand{\argmax}{\operatornamewithlimits{argmax}}

\definecolor{darkblue}{rgb}{0, 0, 0.5}
\hypersetup{colorlinks=true,citecolor=darkblue, linkcolor=darkblue, urlcolor=darkblue}

\issue{1}{1}{2016}

\runningtitle{Neural Metric Learning for Fast End-to-End Relation Extraction}

\runningauthor{Tran et al.}

\begin{document}

\title{Neural Metric Learning for Fast End-to-End Relation Extraction}

\author{Tung Tran}
\affil{Department of Computer Science\\
University of Kentucky, USA\\
\texttt{tung.tran@uky.edu}}

\author{\,}

\author{Ramakanth Kavuluru}
\affil{Division of Biomedical Informatics \\
Department of Internal Medicine 
Department of Computer Science\\
University of Kentucky, USA\\
\texttt{ramakanth.kavuluru@uky.edu}}

\maketitle

\begin{abstract}
  Relation extraction (RE) is an indispensable information extraction task in several disciplines.~RE models typically assume that named entity recognition (NER) is already performed in a previous step by another independent model.~Several recent efforts, under the theme of \emph{end-to-end} RE, seek to exploit inter-task correlations by modeling both NER and RE tasks jointly.~Earlier work in this area commonly reduces the task to a table-filling problem wherein an additional expensive decoding step involving beam search is applied to obtain globally consistent cell labels. In efforts that do not employ table-filling, global optimization in the form of CRFs with Viterbi decoding for the NER component is still necessary for competitive performance.~We introduce a novel neural architecture utilizing the table structure, based on repeated applications of 2D convolutions for pooling local dependency and metric-based features, that improves on the state-of-the-art
  without the need for global optimization.~We validate our model on the ADE and CoNLL04 datasets for end-to-end RE and demonstrate $\approx 1\%$ gain (in F-score) over prior best results with training and testing times that are seven to ten times faster --- the latter highly advantageous for time-sensitive end user applications.
\end{abstract}

\section{Introduction}

Information extraction (IE) systems are fundamental to the automatic construction of knowledge bases and ontologies from unstructured text. While important, in and of themselves, these resulting resources can be harnessed to advance other important language understanding applications including knowledge discovery and question answering systems. Among IE tasks are named entity recognition (NER) and binary relation extraction (RE) which involve identifying named entities and relations among them, respectively, where the latter is typically a set of triplets identifying pairs of related entities and their relation types. 

\begin{figure}[t]
  \center{\includegraphics[width=0.7\columnwidth]
  {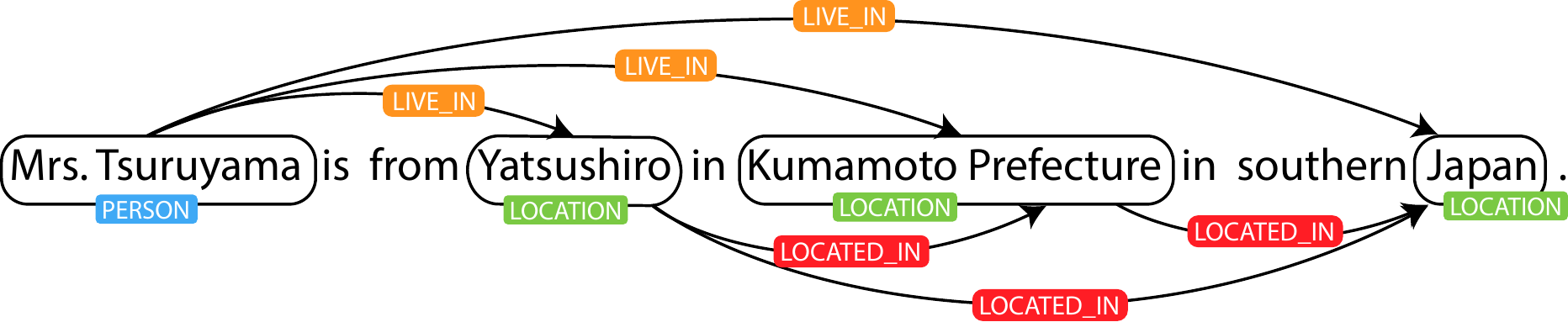}}
  \caption{A simple relation extraction example.}
  \label{fig_example}
\end{figure}

We present Figure~\ref{fig_example} as an example of the NER and RE problem given the input sentence 
``Mrs.~Tsuruyama is from Yatsushiro in Kumamoto Prefecture in southern Japan.'' First, we extract as entities the spans ``Mrs.~Tsuruyama'', ``Yatsushiro'', ``Kumamoto Prefecture'', and ``Japan'' where ``Mrs.~Tsuruyama'' is of type \texttt{PERSON} and the rest are of type \texttt{LOCATION}. Thus, NER consists of identifying both the bounds and type of entities mentioned in the sentence. Once entities are identified, the next step is to extract relation triplets of the form \texttt{(subject,predicate,object)}, if any, based on the context; for example, \texttt{(Mrs. Tsuruyama, LIVE\_IN, Yatsushiro)} is a relation triple that may be extracted from the example sentence as output of an RE system. Given this, it is clear that \ere{} is a complex problem given the sparse nature of the output space; for a sentence of $n$ length with $k$ possible relation types, the output is a variable-length set of relations each drawn from $kn^2$ possible relation combinations.

NER and RE have been traditionally treated as independent problems to be solved separately and later combined in an ad-hoc manner as part of a pipeline system. End-to-end RE (\ere{}) is a relatively new research direction that seeks to model NER and RE jointly in a unified architecture. As these tasks are closely intertwined, joint models that simultaneously extract entities and their relations in a single framework have the capacity to exploit inter-task correlations and dependencies leading to potential performance gains. Moreover, joint approaches, like our method, are better equipped to handle datasets where entity annotations are non-exhaustive (that is, only entities involved in a relation are annotated), since standalone NER systems are not designed to handle incomplete annotations.
Recent advancements in deep learning for \ere{} are broadly divided into two categories:
(1).~The first category involves applying deep learning to the table structure first introduced by \namecite{miwa2014modeling}, including \namecite{gupta2016table}, \namecite{pawar2017end}, and \namecite{zhang2017end} where \ere{} is reduced to some variant of the table-filling problem such that the $(i,j)$-th cell is assigned a label that represents the relation between tokens at positions $i$ and $j$ in the sentence. We further describe the table-filling problem in Section~\ref{sec-table-fill}. 
Recent approaches based on the table structure operate on the idea that cell labels are dependent on features or predictions derived from preceding or adjacent cells; hence, the table is filled incrementally  leading to potential efficiency issues. Also, these methods typically require an additional expensive decoding step, involving beam search, to obtain a globally optimal table-wide label assignment. 
(2).~The second category includes models where NER and RE are modeled jointly with shared components or parameters without the table structure. Even state-of-the-art methods not utilizing the table structure rely on conditional random fields (CRFs) as an integral component of the NER subsystem where   Viterbi algorithm is used to decode the best  label assignment at test time~\cite{bekoulis2018join,bekoulis2018adversarial}. 

Our model utilizes the table formulation by embedding features along the third dimension. We overcome efficiency issues by utilizing a more efficient and effective approach for deep feature aggregation such that local metric, dependency, and position based features are simultaneously pooled --- in a $3 \times 3$ cellular window --- over many applications of the 2D convolution. Intuitively, preliminary decisions are made at earlier layers and corroborated at later layers. Final label assignments for both NER and RE are made simultaneously via a simple \emph{softmax} layer. 
Thus, computationally, our model is expected to improve over earlier efforts without a costly decoding step. We validate our proposed method on the CoNLL04 dataset~\cite{roth2004linear} and the ADE dataset~\cite{gurulingappa2012development}, which correspond to the general English and the biomedical domain respectively, and show that our method improves over   prior state-of-the-art in \ere{}. We also show that our approach leads to training and testing times that are seven to ten times faster, where the latter can be critical for time-sensitive end-user applications. Lastly, we perform extensive error analyses and show that our network is visually interpretable by examining the activity of hidden pooling layers (corresponding to intermediate decisions). To our knowledge, our study is the first to perform this type of visual analysis of a deep neural architecture for end-to-end relation extraction.\footnote{Our code is included as supplementary material and will be made publicly available on GitHub.}

\section{Related Work}

In this section, we provide an overview of three main types of relation extraction methods in the literature: studies that are limited to relation classification, early \ere{} methods that assume known entity bounds, and recent efforts on \ere{} that perform full entity recognition and relation extraction in an end-to-end fashion. 

\subsection{Relation Classification} 
The majority of past and current efforts in relation extraction treat the problem as a simpler \emph{relation classification} problem where pairs of entities are known during test time; the goal is to classify the pair of entities, given the context, as being either positive or negative for a particular type of relation. Many works on relation classification preprocess the input as a dependency parse tree~\cite{bunescu2005shortest,qian2008exploiting} and exploit features corresponding to the shortest dependency path between candidate entities; this general approach has also been successfully applied in the biomedical domain~\cite{airola2008all,fundel2007relex,li2008kernel,ozgur2008identifying}, where they typically involve a graph kernel based Support Vector Machine (SVM) classifier~\cite{li2008kernel,rink2011automatic}. The concept of network centrality has also been explored~\cite{ozgur2008identifying} to extract gene-disease relations. Other studies, such as the effort by \namecite{frunza2011machine}, apply the more traditional \emph{bag-of-words} approach focusing on syntactic and lexical features while exploring a wide variety of classification algorithms including decision trees, SVMs, and Na\"ive Bayes. More recently, innovations in relation extraction have centered around designing meaningful deep learning architectures. \namecite{liu2016dependency} proposed a dependency-based convolutional neural network (CNN) architecture wherein the convolution is applied over words adjacent according to the shortest path connecting the entities in the dependency tree, rather than words adjacent with respect to the order expressed, to detect drug-drug interactions (DDIs). In \namecite{kavuluru2017extracting}, ensembling of both character-level and word-level recurrent neural networks (RNNs) is further proposed for improved performance in DDI extraction. \namecite{raj2017learning} proposed a deep learning architecture such that word representations are first processed by a bidirectional RNN followed by a standard CNN, with an optional attention mechanism towards the output layer. \namecite{zhang2018graph} showed that relation extraction performance can be improved by applying graph convolutions over a pruned version of the dependency tree.

\subsection{End-to-End Relation Extraction with Known Entity Bounds}
Early efforts in \ere{}, as covered in this section, assume that entity bounds are known during test time. Hence, the NER aspect of these methods is limited to classifying entity type (e.g., is "President Kennedy" a person, place, or organization?). In a seminal work, \namecite{roth2004linear} proposed an integer linear programming~(LP) approach to tackle the end-to-end   problem. They discovered that the LP component was effective in enhancing classifier results by reducing semantic inconsistencies in the predictions compared to a traditional pipeline   wherein the outputs of an NER component are passed as features into the RE component. Their results indicate that there are mutual inter-dependencies between NER and RE as subtasks which can be exploited. The LP technique has been also been successfully applied in jointly modeling entities and relations with respect to opinion recognition by  \namecite{choi2006joint}. \namecite{kate2010joint} proposed a similar approach but presented a global inference mechanism induced by building a graph resembling a card-pyramid structure. A dynamic programming algorithm, similar to CYK~\cite{jurafsky2008speech} parsing, called \emph{card-pyramid} parsing is applied along with beam search to identify the most probable joint assignment of entities and their relations based on outputs of local classifiers. Other efforts to this end involve the use of probabilistic graphical models~ \cite{yu2010jointly,singh2013joint}. 

\subsection{End-to-End Relation Extraction}
\namecite{li2014incremental} proposed one of  the first truly joint models wherein entities, including entity mention bounds, and their relations are predicted. Structured perceptrons~\cite{collins2002discriminative}, as a learning framework, are used to estimate feature weights while beam search is used to explore partial solutions to incrementally arrive at the most probable structure. \namecite{miwa2014modeling}  proposed the idea of using a table representation which simplifies the task into a table-filling problem such that NER and relation labels are assigned to cells of the table; the aim was to predict the most probable label assignment to the table, out of all possible assignments, using beam search. While the representation is in table form, beam search is performed sequentially, one cell-assignment per step. The table-filling problem for \ere{} has since been successfully transferred to the deep neural network setting~\cite{gupta2016table,pawar2017end,zhang2017end}.

Other recent approaches not utilizing a table structure involve modeling the entity and relation extraction task jointly with shared parameters~\cite{miwa2016end,li2016joint,zheng2017joint_hybrid,li2017neural,katiyar2017going,bekoulis2018join,zeng2018extracting}. \namecite{katiyar2017going} and \namecite{bekoulis2018join} specifically use attention mechanisms for the RE component without the need for dependency parse features. \namecite{zheng2017joint_tag} operate by reducing the problem to a sequence-labeling task that relies on a novel tagging scheme. \namecite{zeng2018extracting} use an encoder-decoder network such that the input sentence is encoded as fixed-length vector and decoded to relation triples directly. Most recently, \namecite{bekoulis2018adversarial} found that adversarial training (AT) is an effective regularization approach for \ere{} performance.

\section{Method}

We present our version of the table-filling problem, a novel neural network architecture  to fill the table, and details of the training process. Here, Greek letter symbols are used to distinguish hyper-parameters from variables that are learned during training. 

\subsection{The Table-Filling Problem}\label{sec-table-fill}

Given a sentence of length  $n$, we  use an $n \times n$ table to represent a set of semantic relations such that the $(i, j)$-th cell represents the relationship (or non-relation) between tokens $i$ and $j$. In practice, we assign a tag for each cell in the table such that entity tags are encoded along the diagonal while relation tags are encoded at non-diagonal cells. For entity recognition, we use the BILOU tagging scheme~\cite{ratinov2009design}. In the BILOU scheme, \emph{B}, \emph{I}, and \emph{L} tags are used to indicate the beginning, inside, and last token of a multi-token entity respectively. The \emph{O} tag  indicates whether the token outside of an entity span, and \emph{U} is used for unit-length entities. 

\begin{figure}[t]
  \center{\includegraphics[width=0.5\columnwidth]
  {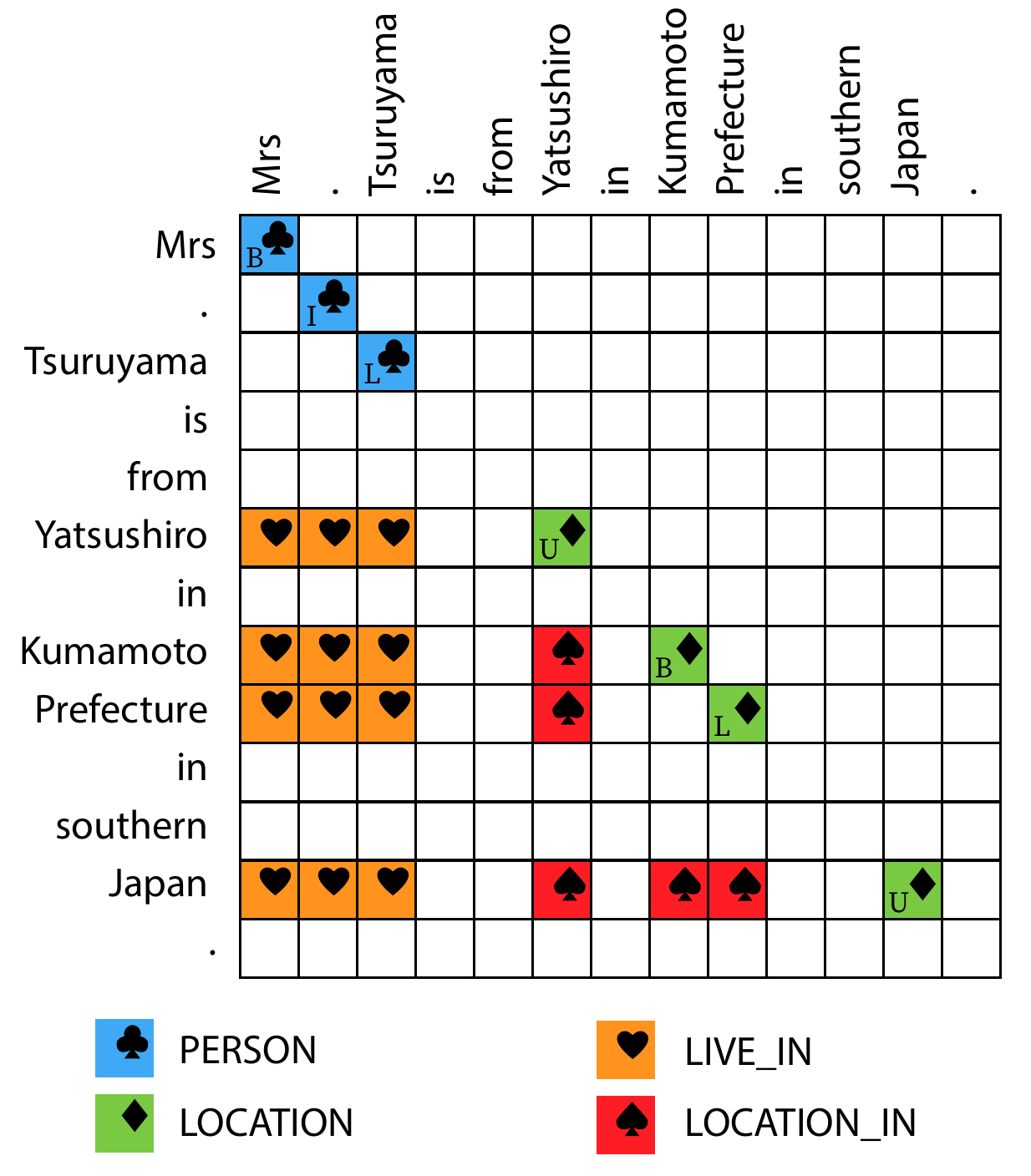}}
  \caption{Table representation for the example in Figure~\ref{fig_example}. BILOU-encoded entity tags are assigned along the diagonal and relation tags are assigned where entity spans intersect. Empty cells are implicitly assigned the O tag.}
  \label{fig_table_rep}
\end{figure}

In tabular form, entity and relation tags are drawn from a unified list $\set{Z}$ serving as the label space; that is, each cell in the table is assigned exactly one tag from $\set{Z}$. For simplicity, the \emph{O} tag is also used to indicate a \emph{null} relation when occurring outside of a diagonal. As each entity type requires a BILOU variant, a problem with $n_{\text{ent}}$ entity types and $n_{\text{rel}}$ relation types has $|\set{Z}| = 4n_{\text{ent}} + n_{\text{rel}} + 1$ where the last term accounts for the \emph{O} tag. Our conception of the table-filling problem differs from \namecite{miwa2014modeling} in that we utilize the entire table as opposed to only the lower triangle; this allows us to model directed relations without the need for additional inverse-relation tags. Moreover, we assign relation tags to cells where entity spans intersect instead of where head words intersect; thus encoded relations manifest as rectangular blocks in the proposed table representation. We present a visualization of our table representation in Figure~\ref{fig_table_rep}. At test time, entities are first extracted, and relations are subsequently extracted by averaging the output probability estimates of the blocks where entities intersect. We describe the exact procedure for extracting relations from these blocks at test-time in Section~\ref{sec-decode}.

\subsection{Our Model: Relation-Metric Network}

\begin{figure*}[t]
  \center{\includegraphics[width=\columnwidth]
  {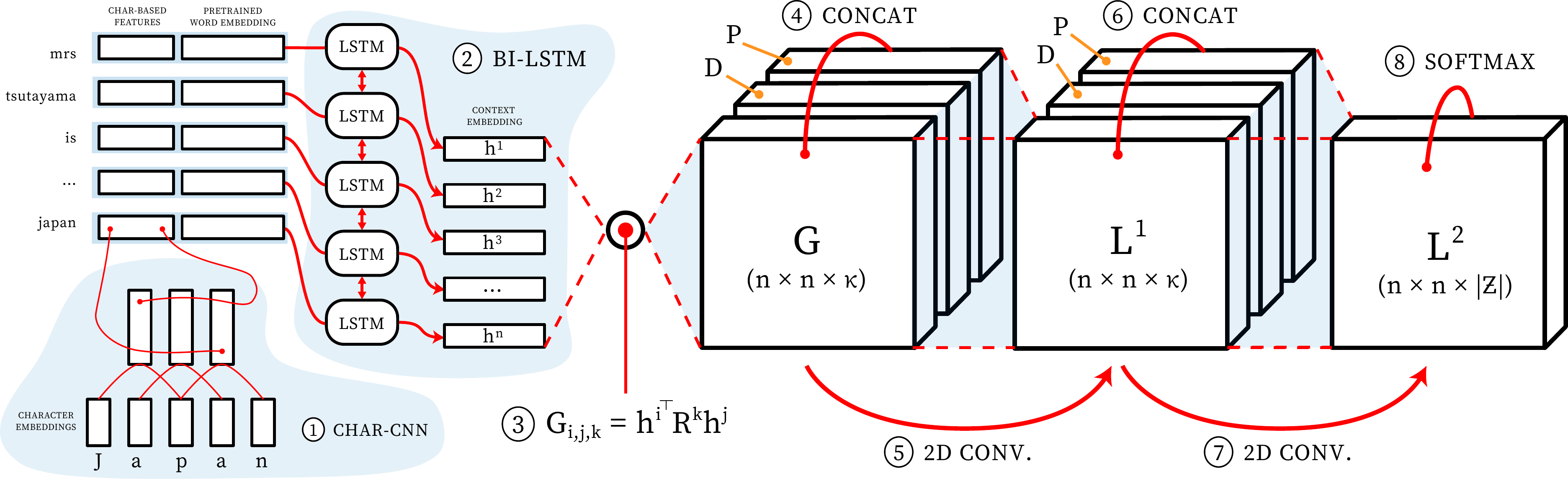}}
  \caption{Overview of the network architecture for $\lambda = 2$. For simplicity, we ignore punctuation tokens.}
  \label{fig_network}
\end{figure*}

We propose a novel neural architecture, which we call the relation-metric network, combining the ideas of metric learning and convolutional neural networks (CNNs) for table filling. The schematic of the network is shown in Figure~\ref{fig_network}, whose components will be detailed in this section. 

\subsubsection{Context Embeddings Layer}

In addition to word embeddings, we employ character-CNN based representations as commonly observed in recent neural NER models~\cite{chiu2016named} and \ere{} models~\cite{li2017neural}. Character-based features can capture morphological features and help generalize to out-of-vocabulary words. 
For the proposed model, such representations are composed by convolving over character embeddings of size $\pi$ using a window of size 3, producing $\eta$ feature maps; the feature maps are then max-pooled to produce $\eta$-length feature representations. As our approach is standard, we refer readers to \namecite{chiu2016named} for full details. This portion of the network is illustrated in step \circled{1} of Figure~\ref{fig_network}.

Suppose the input is a sentence of length $n$ represented by a sequence of word indices $w_1, \ldots, w_n$ into the vocabulary $\set{V}^{\text{Word}}$. Each word is mapped to an embedding vector via embedding matrices $E^{\text{Word}} \in \RR^{|\set{V}^{\text{Word}}| \times \delta}$ such that $\delta$ is a hyperparameter that determines the size of word  embeddings. Next, let $\matrow{C}{i}$ be the character-based representation for the \nth{i} word. An input sentence is represented by matrix $S$ wherein rows are words mapped to their corresponding embedding vectors; or concretely,  
\begin{equation*}\label{eq_input}
S = \left(
\begin{array}{c}
\matrow{E^{\text{Word}}}{w_1} \concat \matrow{C}{1} \\
\vdots\\
\matrow{E^{\text{Word}}}{w_n} \concat \matrow{C}{n} \\
\end{array}
\right) \\
\end{equation*}
where $\concat$ is the vector concatenation operator and $\matrow{E^{\text{Word}}}{i}$ is the \nth{i} row of $E^{\text{Word}}$. 

Next, we compose \emph{context} embedding vectors~(CVs), which embed each word of the sentence with additional contextual features. Suppose $\LSTMR$ and $ \LSTML$ represent a long short term memory (LSTM) network composition in the forward and backward direction, respectively, and let $\rho$ be a hyperparameter that determines context embedding size. We feed $S$ to a Bi-LSTM layer of hidden unit size $\frac{1}{2} \rho$ such that 
\begin{align*}
  \lvec{h}^i &= \LSTMR (\matrow{S}{i}), \,\,\\
  \rvec{h}^i &= \LSTML (\matrow{S}{i}), \\
  \vec{h}^i &= \lvec{h}^i  \concat \rvec{h}^i,
  \text{ \hspace{1em} for } i = 1,\ldots,n,
\end{align*} 
where $\matrow{S}{i}$ is the \nth{i} row of S and $\vec{h}^i \in \RR^{\rho}$ represents the context centered at the \nth{i} word. The output of the Bi-LSTM can be represented as a matrix $H \in \RR^{n \times \rho}$ such that $H = \left( \vec{h}^1, \ldots, \vec{h}^n \right)^{\top}$. This concludes step \circled{2} of Figure~\ref{fig_network}.

\subsubsection{Relation-Metric Learning}

Our goal is to design a network such that any two CVs can be compared via some ``relatedness'' measure; that is, we wish to learn a relatedness measure (as a parameterized function)  that is able to capture correlative features indicating semantic relationships. A common approach in metric learning to parameterize a relatedness function is to model it in bilinear form. Here, for input vectors $\vec{x}, \vec{z} \in \RR^{m}$, a similarity function in bilinear form is formally defined as  
\begin{equation}\label{eq-bilinear-sim}
  s_R(\vec{x},\vec{z}) = \vec{x}^{\top}R\vec{z}
\end{equation}
where $R \in \RR^{m \times m}$ is a parameter of the relatedness function, dubbed a \emph{relation-metric embedding} matrix, that is learned during the training process. 

In machine learning research, Eq.~\ref{eq-bilinear-sim} is also associated with a type of attention mechanism commonly referred to as ``multiplicative'' attention~\cite{luong2015effective}. However, we apply Eq.~\ref{eq-bilinear-sim} with the classical goal of learning a variety of metric-based features. 
Our aim is to compute $s_R$ for all pairs of CVs in the sentence. Concretely, we can compute a ``relational-metric table''  $G \in \RR^{n \times n}$ over all pairs of CVs in the sentence such that $G_{i,j} = {\vec{h}^{i}}^{\top} R \vec{h}^j$.
In fact, we can learn a collection of $\kappa$ similarity functions corresponding to $\kappa$ relation metric tables; for our purposes, this is analogous to learning a diverse set of convolution filters in the context of CNNs. Thus we have the 3-dimensional tensor
\begin{equation}\label{eq-metric}
  G_{i,j,k} = {\vec{h}^{i}}^{\top} R^k \vec{h}^j, \text{ \hspace{1em} for } k = 1,\ldots,\kappa,
\end{equation}
with $G \in \RR^{n \times n \times \kappa}$ where the first and second dimension correspond to word position indices while the third dimension embeds metric-based features.
This constitutes step \circled{3} of Figure~\ref{fig_network}. We show how $G$ is consumed by the rest of the network in Section~\ref{sec-pooling}. However, as a prerequisite, we first describe how dependency parse and relative position information is prepared in Section~\ref{sec-dep} and Section~\ref{sec-pos} respectively and define the 2D convolution in Section~\ref{sec-cnn}.

\subsubsection{Dependency Embeddings Table}\label{sec-dep}

Let $\set{V}^{\text{dep}}$ be the vocabulary of syntactic dependency tags (e.g., \texttt{nsubj, dobj}).~For an input sentence, let $\set{T} = \{ (a_1,b_1,z_1), \ldots , (a_{\hat{d}},b_{\hat{d}},z_{\hat{d}}) \}$ be the set of dependency relations where  $z_i$ are mappings to tags in $\set{V}^{\text{dep}}$ that express the dependency-based relations between  pairs of words at positions $a_i, b_i \in \{ 1, \ldots, n\}$, respectively. We define the dependency embedding matrix as $F^{\text{dep}} \in \RR^{|\set{V}^{\text{dep}}| \times \beta}$, where each unique dependency tag is a $\beta$-dimensional embedding. We compose the dependency representation tensor $D$ for $\set{T}$ as
\begin{equation*}\label{eq-dep}
  D_{i,j,k} = \begin{cases}
  \,\, F^{\text{dep}}_{t,k}  & \text{if} \,\, (i,j,t) \in \set{T} \,\, \text{or} \,\, (j,i,t) \in \set{T},\\
  \,\, \vec{\phi}_k & \text{otherwise},
  \end{cases}
\end{equation*}
for $k = 1,\ldots,\beta$, where $\vec{\phi}$ is a trainable embedding vector representing the \emph{null} dependency relation. As shown in the above equation for $ D_{i,j,k}$, we embed the dependency parse tree simply as an undirected graph. 

\subsubsection{Position Embeddings Table}\label{sec-pos} 

First proposed by \namecite{zeng2014relation}, so called \emph{position vectors} have been shown to be effective in neural models for relation classification. Position vectors are designed to encode the relative offset between a word and the two candidate entities (for RE) as fixed-length embeddings. We bring this idea to the tabular setting by proposing a \emph{position} embeddings table $P$, which is composed  the same way as the dependencies table; however, instead of  dependency tags, we simply encode the distance between two candidate CVs as discrete labels mapped to fixed-length embeddings (of size $\gamma$, a hyperparameter). It is straightforward to see there will be $2(n_{\text{max}} -1)+1$ distinct position offset labels where $n_{\text{max}}$ is the maximum length of a sentence in the training data. Specifically, given a position vocabulary $\set{V}^{\text{dist}}$, associated position embedding matrix $F^{\text{dist}} \in \RR^{|\set{V}^{\text{dist}}| \times \gamma}$, the position embeddings tensor is $P_{i,j,k} = F^{\text{dist}}_{(i-j),k}$ for $k = 1,\ldots,\gamma$. As an implementation detail, we set $\set{V}^{\text{dist}}$ to  $\{ -n_{\text{max}},\ldots,n_{\text{max}} \}$ where $n_{\text{max}}$ is the maximum sentence length over all training examples. 
Both   dependency and position embedding tensors are concatenated to the metric tensor (Eq.~\eqref{eq-metric}) along the 3rd dimension prior to every convolution operation. Hence they are shown in steps \circled{4} and \circled{6} of Figure~\ref{fig_network} for the network with two convolutional layers. 

\subsubsection{2D Convolution Operation}\label{sec-cnn}

Unlike the standard 2D convolution typically used in NLP tasks, which takes 2D input, our 2D convolution operates on 3D input commonly seen in computer vision tasks where colored image data has height, width, and an additional dimension for color channel. The goal of the 2D convolution is to pool information within a $3 \times 3$ window along the first two dimensions such that metric features and dependency/positional information of adjacent cells are pooled locally over several layers. However, it is necessary to perform a \emph{padded} convolution to ensure that dimensions corresponding to word positions are not altered by the convolution. We denote this padding transformation using the \emph{hat} accent. That is, for some tensor input $X \in \RR^{n \times n \times m}$, the padded version is $\widehat{X} \in \RR^{(n+2)  \times (n+2) \times m}$ and the zero-padding exists at the beginning and at the end of the first and second dimensions. Next, we define the 2D convolution operation via the  $\star$ operator which corresponds to an element-wise product of two tensors followed by summation over the products; formally, for two input tensors $A$ and $B$, $A \star B = \sum_i \sum_j \sum_k A_{i,j,k} B_{i,j,k}$.

\begin{figure}[t]
  \center{\includegraphics[width=0.8\columnwidth]
  {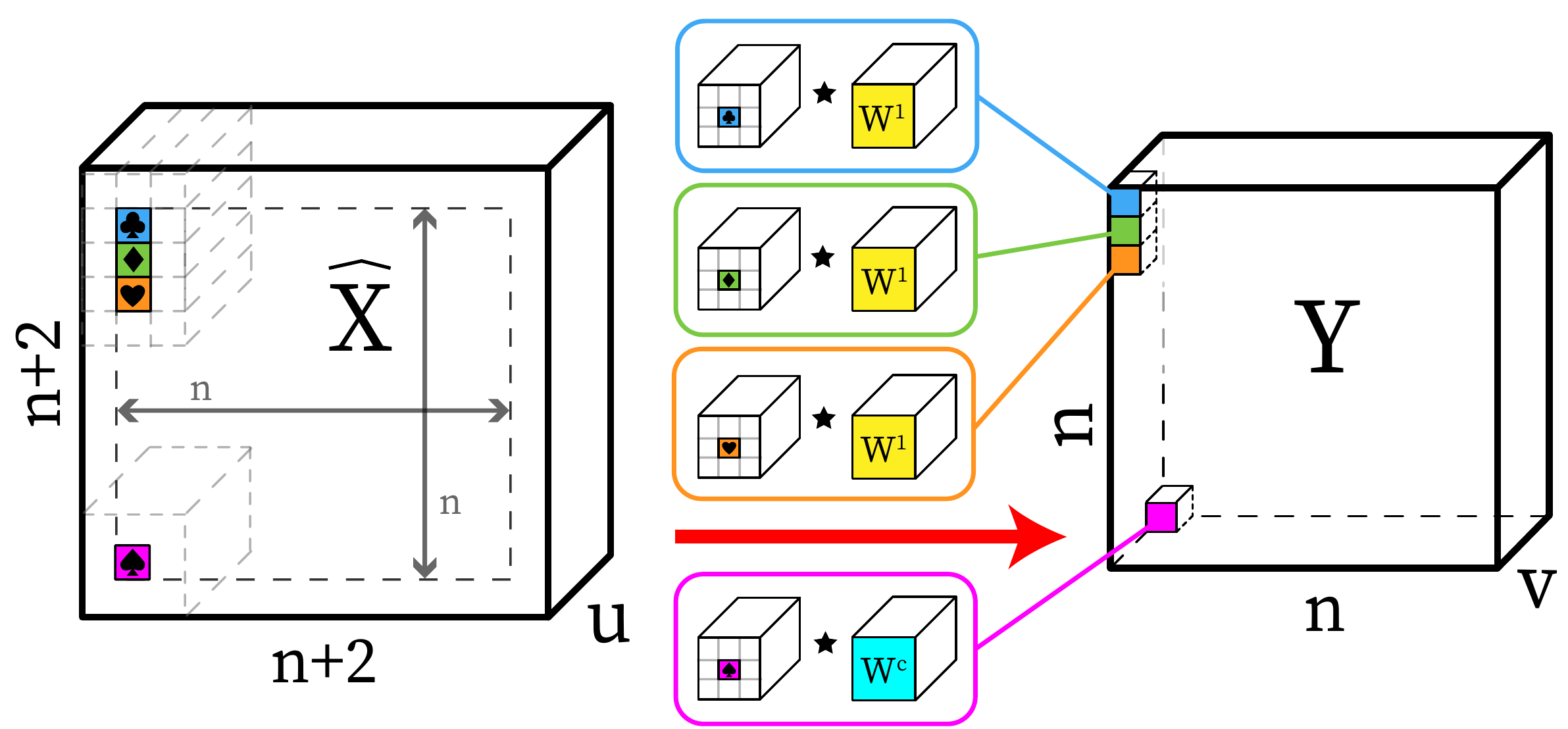}}
  \caption{2D convolution on 3D input with padding}
  \label{fig_2d_conv}
\end{figure}

Now our 2D convolution step is a tensor map $f_v(X): \RR^{n \times n \times u} \to \RR^{n \times n \times v}$  with $v$ filters of size $3 \times 3 \times u$, defined as  
\begin{equation}\label{eq-conv}
  f_v(X)_{i,j,k} = W^k \star \widehat{X}_{[i:i+2][j:j+2][1:u]} + \vec{b}_k
\end{equation}
for $i = 1,\ldots,n, j = 1,\ldots,n, k = 1,\ldots,v,$ where $W^k \in \RR^{3 \times 3 \times u}$ for $k = 1,\ldots,v$, and $\vec{b} \in \RR^{v}$ are filter and bias variables respectively, and $\widehat{X}_{[i:i+2][j:j+2][1:u]}$ is a $3 \times 3 \times u$ window of $\widehat{X}$ from $i$ to $i+2$ along the first dimension, $j$ to $j+2$ along the second dimension, and $1$ to $u$ along the final dimension. We show how $f_v(X)$ is used to repeatedly pool contextual information in Section~\ref{sec-pool}. Instead of a $3 \times 3$ window, the convolution operation can be over any  $t \times t$ window for some odd $t \geq 3$ where large $t$ values lead to larger parameter spaces and multiplication operations. The 2D convolution is illustrated in Figure~\ref{fig_2d_conv} and manifests in steps \circled{5} and \circled{7} of Figure~\ref{fig_network}. 

\subsubsection{Pooling Mechanism}\label{sec-pooling}
\label{sec-pool}
Central to our architecture is the iterative \emph{pooling} mechanism designed so that preliminary decisions are made in early iterations and further corroborated in subsequent iterations. It also facilitates the propagation of local metric and dependency/positional features to neighboring cells. Let $\set{Z}$ be the set of tags for the target task. We denote hyper-parameters $\kappa$ and $\lambda$ as the number of channels and the number of CNN layers respectively, where $\kappa$ is same hyperparameter previously defined to represent the size of metric-based features. 
 The pooling layers are defined recursively with base case $L^1 = \text{relu} ( f_{\kappa}(G \concat D \concat P ) )$ and
\begin{equation*}
  L^i = \begin{cases}
     \,\, \text{relu} ( f_{\kappa}(L^{i-1} \concat D \concat P ) ) & \,\, 1 < i < \lambda,\\
    \,\, f_{|\set{Z}|}( L^{i-1} \concat D \concat P ) & \,\, i = \lambda,\\
    \end{cases}
\end{equation*}
where $f$ is the convolution function from Eq.~\eqref{eq-conv}, $G$ is the tensor from Eq.~\eqref{eq-metric}, and $\concat$ is the tensor concatenation operator along the third dimension, and $\text{relu}(x) = \max (0,x)$ is the linear rectifier activation function. Here, $\kappa$ and $\lambda$ determine the breadth and depth of the architecture. A higher $\lambda$  corresponds to a larger receptive field when making final predictions. For example at $\lambda = 2$, the decision at some cell is informed by its immediate neighbors with a receptive field of $3 \times 3$. However, at $\lambda = 3$, decisions are informed by all adjacent neighbors in a $5 \times 5$ window. 
The last layer, $L^{\lambda}$, is the output layer immediately prior to application of the \textit{softmax} function.
Given the architecture in Figure~\ref{fig_network} with two convolutional layers, the convolve-and-pool operation is applied twice, indicated as steps  \circled{5} and \circled{7} in the figure. 

\subsubsection{Softmax Output Layer}

Given $L^{\lambda}$, we apply the \textit{softmax} function along the third dimension to obtain a categorical distribution tensor $Q \in \RR^{n \times n \times |\set{Z}|}$over output tags $\set{Z}$ for each word position pair such that $Q_{i,j,k} = \exp (L^{\lambda}_{i,j,k})/(\sum_{l=1}^{|\set{Z}|}\exp (L^{\lambda}_{i,j,l}))$, where $Q_{i,j,k}$ is the probability estimate of the pair of words at position $i$ and $j$ being assigned the $k$th tag. This constitutes the final step \circled{8} of the network (Figure~\ref{fig_network}). 
Suppose $Y \in \RR^{n \times n \times |\set{Z}|}$ represents the corresponding one-hot encoded  ground truth  along the third dimension such that $Y_{i,j,k} \in \{0,1\}$. Then the example-based loss $\ell$ is obtained by summing the categorical cross-entropy loss over each cell in the table, normalized by the number of words in the sentence; that is, 
\begin{equation*} 
\ell (Y,Q; \theta) = - \frac{1}{n} \sum_{i=1}^n \sum_{j=1}^n \sum_{k=1}^{|\set{Z}|} Y_{i,j,k} \log(Q_{i,j,k}),
\end{equation*}
where $\theta$ is the network parameter set. During training, the loss $\ell$ is computed per example and averaged along the mini-batch dimension.

\subsection{Decoding}\label{sec-decode}

While we learn concrete tags during training, the process for extracting predictions is slightly more nuanced. Entity spans are straightforwardly extracted by decoding BILOU tags along the diagonal. However, RE is based on ``ensembling'' the cellular outputs of the table where entity spans intersect. For entities $a$ and $b$ represented by their starting and ending offsets, $(a_{\text{S}}, a_{\text{E}})$ and $(b_{\text{S}}, b_{\text{E}})$, the relation between them is the label computed as  
$
\argmax_{1 \leq k \leq |\set{Z}|} \,\, \sum_{i = a_{\text{S}}}^{a_{\text{E}}} \sum_{j = b_{\text{S}}}^{b_{\text{E}}} Q_{i,j,k},
$
which indexes a tag in the label space $\set{Z}$.

\section{Experimental Setup}

In this section, we describe the established evaluation method, the datasets used for training and testing, and the configuration of our model. We note that the computing hardware is controlled across experiments given we  report training and testing run times. Specifically, we used the Amazon AWS EC2 \texttt{p2.xlarge} instance which supports the NVIDIA Tesla K80 GPU with 12 GB memory.

\subsection{Evaluation Metrics}

We use the well-known F1 measure (along with precision and recall) to evaluate NER and RE subtasks as in prior work. For NER, a predicted entity is treated as a \emph{true positive} if it is exactly matched to an entity in the groundtruth based on both character offsets and entity type. For RE, a predicted relation is treated as a \emph{true positive} if it is exactly matched to a relation in the ground truth based on subject/object entities and relation type. As relation extraction performance directly subsumes NER performance, we focus purely on relation extraction performance as the primary evaluation metric of this study.

\subsection{Datasets}


\paragraph{CoNLL04} We use the dataset originally released by \namecite{roth2004linear} with 1441 examples consisting of news articles from outlets such as WSJ and AP. The dataset has four entity types including \emph{Person}, \emph{Location}, {Organization}, and \emph{Other} and five relation types including \emph{Live\_In}, \emph{Located\_In}, \emph{OrgBased\_In}, \emph{Work\_For}, and \emph{Kill}. We report results based on training/testing on the same train-test split as established by \namecite{gupta2016table,adel2017global,bekoulis2018join,bekoulis2018adversarial}, which consists of 910 training, 243 development, and 288 testing instances.

\paragraph{ADE} We also validate our method on the Adverse Drug Events (ADE) dataset from \namecite{gurulingappa2012development} for extracting drug-related adverse effects from medical text. Here, the only entity types are \emph{Drug} and \emph{Disease} and the relation extraction task is strictly binary (i.e., Yes/No w.r.t the ADE relation). The examples come from 1644 PubMed abstracts and are divided in two partitions: the first partition of 6821 sentences contain at least one drug/disease pair while the second partition of 16695 sentences contain no drug/disease pairs. As with prior work~\cite{li2016joint,li2017neural,bekoulis2018join,bekoulis2018adversarial}, we  only use examples from the first partition from which  120 relations with nested entity annotations (such as ``lithium intoxication'' where \emph{lithium} and \emph{lithium intoxication} are the drug/disease pair) are removed. 
Since sentences are duplicated for each pair of drug/disease mention in the original dataset, when collapsed on \emph{unique} sentences, the final dataset used in our experiments constitutes 4271 sentences in total. 
Given there are no official train-test splits, we report results based on 10-fold cross-validation, where results are based on averaging performance across the ten folds, as in prior work.

\subsection{Model Configuration}
 
  \begin{table}[ht]
    \caption{Model configuration as tuned on the CoNLL04 development set.}
    \label{tab_config}
    \centering
    \resizebox{\textwidth}{!}{
    \footnotesize
    \begin{tabular}{cc}
    \toprule
    \begin{tabular}{lc}
    \textbf{Setting} & \textbf{Value}\\
    \midrule
    Optimization Method \hspace{1em} & RMSProp\\
    Learning Rate & 0.005 \\
    Dropout Rate & 0.5\\
    Num. Epochs & 100\\
    Num. Channels ($\kappa$) & 15\\
    Num. Layers ($\lambda$) & 8\\
    \end{tabular} & 
    \begin{tabular}{lc}
    \textbf{Setting} & \textbf{Value}\\
    \midrule
    Character Embedding Size ($\pi$) \hspace{1em} & 25\\ 
    Character Representation Size ($\eta$) & 50\\ 
    Position Embedding Size ($\gamma$) & 25\\ 
    Dependency Embedding Size ($\beta$) & 10\\ 
    Word Embedding Size ($\delta$) & 200\\
    Context Embedding Size ($\rho$) & 200\\
    \end{tabular}\\
    \bottomrule
    \end{tabular}}
    \end{table}

We tuned our model on the CoNLL04 development set; the corresponding configuration of our model (including hyperparameter values) used in our main experiments is shown in  Table~\ref{tab_config}. For the ADE dataset, we used Word2Vec embeddings pretrained on the corpus of PubMed abstracts~\cite{pyysalo2013distributional}. For the CoNLL04 dataset, we used GloVe  embeddings pretrained on Wikipedia and Gigaword~\cite{pennington2014glove}. All other variables are initialized using values drawn from a normal distribution with a mean of $0$ and standard deviation of $0.1$ and further tuned during training. Words were tokenized on both spaces and punctuations; punctuation tokens were kept as is common practice for NER systems. For part-of-speech and dependency parsing, we use the well-known tool  
spaCy\footnote{https://spacy.io/}. For both datasets, we used projective dependency parses produced from the default pretrained English models. We found that using models pretrained on biomedical text (namely, the GENIA~\cite{kim2003genia} corpus) did not improve performance on the ADE dataset.

Early experiments showed that applying exponential decay to the learning rate in conjunction with batch normalization~\cite{ioffe2015batch} is essential for stable/effective learning for this particular architecture.  We apply exponential decay to the learning rate such that it is roughly halved  every 10 epochs; concretely, $r_k = {r_b}^{\frac{k}{10}}$ where $r_b$ is the base learning rate and $r_k$ is the rate at the $k$th epoch. We apply dropout~\cite{srivastava2014dropout} on $h_i$ for $i = 1,\ldots,n$ as regularization at the earlier layers. However, dropout had a detrimental impact when applied to later layers. We instead apply batch normalization as a form of regularization on representations $G$ and $L^i$ for $i = 1,\ldots,\lambda-1$. We  optimize the objective loss using RMSProp~\cite{tieleman2012lecture} with a relatively high initial learning rate of 0.005 given exponential decay is used. 


\section{Results and Discussion}

\begin{table*}[ht]
  \caption{Results comparing to other methods on the CoNLL04 dataset. We report 95\% confidence intervals around the mean F1 over 30 runs for models in the last two rows.  Our model was tuned on the CoNLL04 development set corresponding to the configuration from Table~\ref{tab_config}.}
  \label{tab_results_conll04}
  \renewcommand{\arraystretch}{1.75}
  \resizebox{\textwidth}{!}{
  \begin{tabular}{@{\extracolsep{0.1em}}l ccc ccc cc}
    \toprule
     \, & \multicolumn{3}{c}{Entity Recognition} & \multicolumn{3}{c}{Relation Extraction} & \footnotesize Avg. Epoch & \footnotesize Avg. \\
    \cline{2-4}\cline{5-7}
     \textbf{Model} & \textbf{P} (\%) & \textbf{R} (\%) & \textbf{F} (\%) & \textbf{P} (\%) & \textbf{R} (\%) & \textbf{F} (\%) & \footnotesize Train Time & \footnotesize Test Time $\ast$\\
     \midrule
     Table Representation\cite{miwa2014modeling}	& 81.20	& 80.20	& 80.70 & 76.00 & 50.90 & 61.00 & - & -\\
     Multihead~\cite{bekoulis2018join} & 83.75 & 84.06 & 83.90 & 63.75 & 60.43 & 62.04 & - & -\\
     Multihead with \textbf{AT}~\cite{bekoulis2018adversarial} & - & - & 83.61 & - & - & 61.95 & - & -\\
     \midrule
     Replicating Multihead with \textbf{AT}~\cite{bekoulis2018adversarial}$\dagger$ & 84.36 & 85.80 & \textbf{85.07} \ci{0.26} & 65.81 & 57.59 & 61.38 \ci{0.50} & 614 sec & 34 sec\\
     Relation-Metric (Ours)$\dagger$ & 84.46 & 84.67 & 84.57 \ci{0.29} & 67.97 & 58.18 & \textbf{62.68} \ci{0.46} & \textbf{101 sec} & \textbf{4.5 sec}\\
     \bottomrule
  \end{tabular}}\vspace{0.2em}\\
  \scriptsize{$\dagger$ These results are directly comparable given the same train-test splits, pretrained word embeddings, and computing hardware.}\vspace{0.2em}\\
  \scriptsize{$\ast$ Average test time is per test set of 288 examples; dependency parsing accounts for approximately 0.5 second of our reported test time.}
\end{table*}

\begin{table*}[ht]
  \caption{Results comparing to other methods on the ADE dataset. We report the mean performance over 10-fold cross-validation for models in the last two rows. Our model was tuned on the CoNLL04 development set corresponding to the configuration from Table~\ref{tab_config}.}
    \label{tab_results_ade}
  \renewcommand{\arraystretch}{1.75}
  \resizebox{\textwidth}{!}{
  \begin{tabular}{@{\extracolsep{0.1em}}l ccc ccc cc}
    \toprule
     \, & \multicolumn{3}{c}{Entity Recognition} & \multicolumn{3}{c}{Relation Extraction} & \footnotesize Avg. Epoch & \footnotesize Avg. \\
    \cline{2-4}\cline{5-7}
     \textbf{Model} & \textbf{P} (\%) & \textbf{R} (\%) & \textbf{F} (\%) & \textbf{P} (\%) & \textbf{R} (\%) & \textbf{F} (\%) & \footnotesize Train Time & \footnotesize Test Time $\ast$\\
     \midrule
     Neural Joint Model~\cite{li2016joint} & 79.50 & 79.60 & 79.50 & 64.00 & 62.90 & 63.40 & - & -\\
     Neural Joint Model~\cite{li2017neural} & 82.70 & 86.70 & 84.60 & 67.50 & 75.80 & 71.40 & - & -\\
     Multihead~\cite{bekoulis2018join} & 84.72 & 88.16 & 86.40 & 72.10 & 77.24 &  74.58 & - & -\\
     Multihead with \textbf{AT}~\cite{bekoulis2018adversarial} & - & - & 86.73 & - & - & 75.52 & - & -\\
     \midrule
     Replicating Multihead with \textbf{AT}~\cite{bekoulis2018adversarial}$\dagger$ & 85.76 & 88.17 & 86.95 & 74.43 & 78.45 & 76.36 & 1567 sec & 40 sec\\
     Relation-Metric (Ours)$\dagger$ & 86.16 & 88.08 & \textbf{87.11} & 77.36 & 77.25 & \textbf{77.29} & \textbf{134 sec} & \textbf{4.5 sec}\\
     \bottomrule
  \end{tabular}}\vspace{0.2em}\\
  \scriptsize{$\dagger$ These results are directly comparable given the same fixed 10-fold splits, pretrained word embeddings, and computing hardware.}\vspace{0.2em}\\
  \scriptsize{$\ast$ Average test time is per test set of 427 examples; dependency parsing accounts for approximately 0.5 second of our reported test time.}
\end{table*}

We report our main results in Tables~\ref{tab_results_conll04} and \ref{tab_results_ade} for the CoNLL04 and ADE datasets respectively. As a baseline, we replicate the prior best models~\cite{bekoulis2018adversarial} for both datasets based on publicly available source code\footnote{https://github.com/bekou/multihead\_joint\_entity\_relation\_extraction}. Unlike prior work, which reports performance based on a single run, we report the 95\% confidence interval around the mean F1 based on 30 runs with differing seed values for the CoNLL04 dataset. For the ADE dataset, we instead report the mean performance over 10-fold cross-validation so that results are comparable to established work. These experiments were performed using the same splits, pretrained embeddings, and computing hardware; hence, results are directly comparable. 

We make the following observations based on our results from Table~\ref{tab_results_conll04}. Both our model and the model from \namecite{bekoulis2018adversarial} tend to skew heavily towards precision. However, our method improves on both precision and recall, and by over 1\% F1 on relation extraction where improvements are statistically significant ($p < 0.05$) based on the two-tailed Student's t-test. We note that our model performs slightly worse when evaluated \emph{purely} on NER. We contend this is a worthwhile trade-off given our model is tuned purely on relation extraction and the relation extraction metric, being end-to-end, indirectly accounts for NER performance. Based on Table~\ref{tab_results_ade}, when tested on the ADE dataset, our method improves over prior best results by approximately 1\% F1 for RE on average. While the prior best skews toward recall in this case, our method exhibits better balance of precision and recall.  Based on run time results, we contend that our method is more computationally efficient given training and testing times are nearly seven  times lower on the CoNLL04 and ten times lower on the ADE set when compared to prior efforts. We note that dependency parsing accounts approximately one-half second of our testing time. While training time may not be crucial in most settings, we argue that fast and efficient predictions are important for many end-user applications.

As an auxiliary experiment, we tested the potential for integrating adversarial training (AT) with our model; however, there were no performance gains even with extensive tuning. On the CoNLL04 dataset, our method with AT performs at 62.26\% F1, compared to 62.68\% without AT. On the ADE dataset, our method performs at 76.83\% F1 with AT, compared to 77.29\% without AT. Given this, we have elected not to include AT evaluations as part of our main results.

\paragraph{Comparison with More Prior Efforts} \namecite{gupta2016table}, \namecite{adel2017global}, and \namecite{zhang2017end} also experimented with the CoNLL04 dataset; however, \namecite{gupta2016table} evaluate on a more relaxed evaluation metric for matching entity bounds while \namecite{adel2017global} assume entity bounds are known at test time thus treating the NER aspect  as a simpler entity \emph{classification} problem. Of the three studies, results from \namecite{zhang2017end} are most comparable given they consider entity bounds in their evaluations; however, their results are based on a \emph{random} 80\%--20\% split of the train and test set. As we use established splits based on prior work, the two results are not directly comparable.

\subsection{Ablation Analysis}

\begin{table}[ht]
  \caption{Ablation studies for relation extraction over the CoNLL04 and ADE dataset; each row after the first indicates removal of a particular feature/component.}
  \label{tab_ablation}
  \centering
  \renewcommand{\arraystretch}{1.2}
  \resizebox{\columnwidth}{!}{
  \begin{tabular}{@{\extracolsep{1em}}l ccc ccc ccc}
    \toprule
     \, & \multicolumn{3}{c}{CoNLL04 (Relation)} & \multicolumn{3}{c}{ADE (Relation)}\\
    \cline{2-4} \cline{5-7}
     \textbf{Model} & \textbf{P} (\%) & \textbf{R} (\%) & \textbf{F} (\%) & \textbf{P} (\%) & \textbf{R} (\%) & \textbf{F} (\%)\\
     \midrule     
     Full model & 67.97 & 58.18 & \textbf{62.68} & 77.36 & 77.25 & \textbf{77.29}\\
     -- Character-based Input & 67.30 & 52.69 & 59.09 & 76.73 & 76.44 & 76.58\\
     -- Dependency Embeddings & 66.56 & 57.69 & 61.78 & 75.79 & 77.16 & 76.45\\
     -- Position Embeddings & 68.57 & 57.34 & 62.43  & 75.94 & 76.62 &76.27\\
     -- Pretrained Word Embeddings & 62.33 & 46.09 & 52.96 & 72.50 & 71.41 & 71.91\\
     \bottomrule
  \end{tabular}}
\end{table}

We report  ablation analysis results in Table~\ref{tab_ablation} using our best model as the baseline. We note that the model hyperparameters were tuned on the CoNLL04 development set. Character and dependency based features all had a notable impact on performance for either dataset. On the hand, while position embeddings had a positive effect on the ADE dataset, performance gains were negligible when testing on  CoNLL04. For the CoNLL04 dataset, we find that character based features had little effect on precision while improving recall substantially. 

\begin{figure}[bt]
  \center{\includegraphics[width=0.5\columnwidth,trim={0 0 1cm 1cm},clip]
  {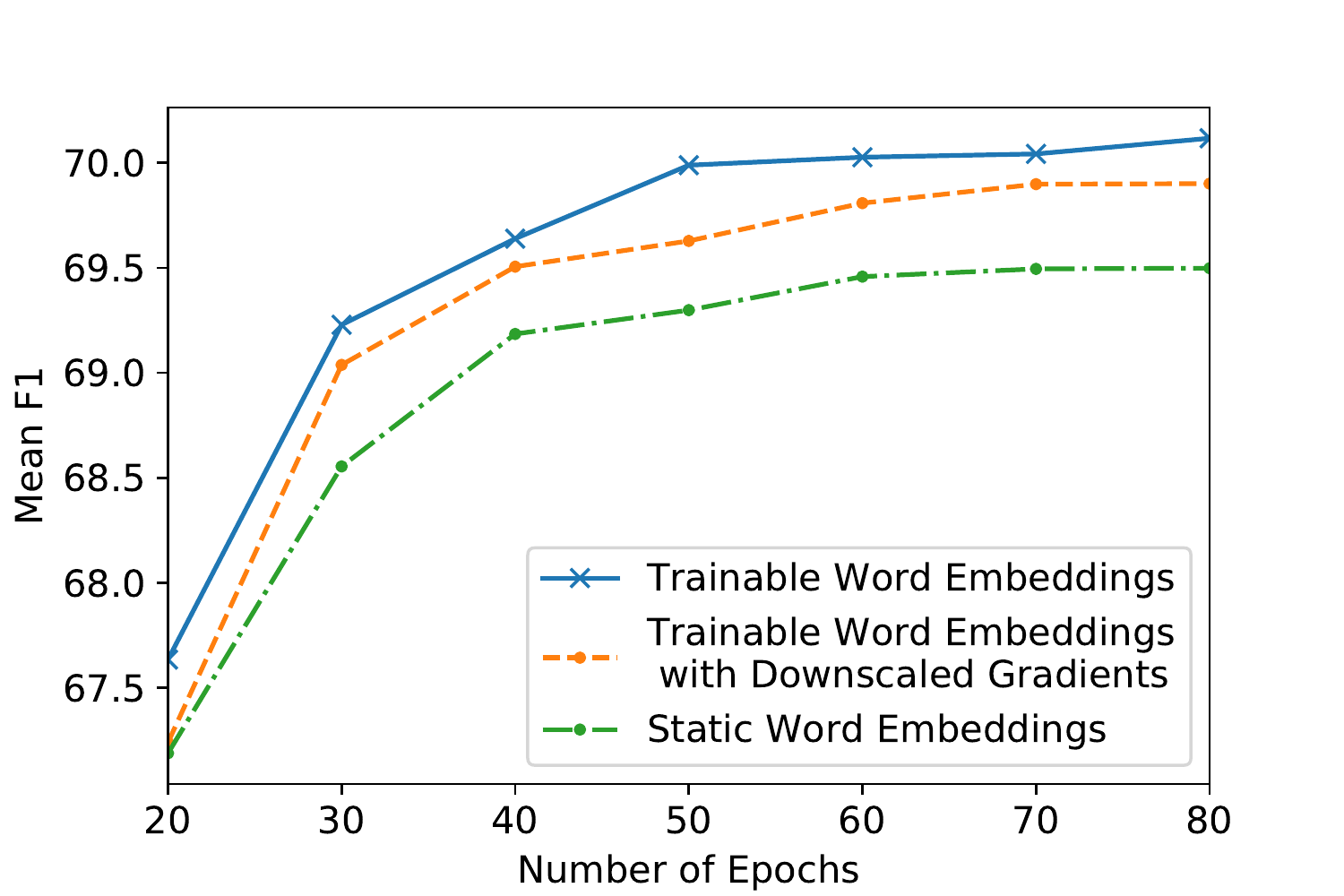}}
  \caption{Mean F1-score (over 10 runs) on CoNLL04 development set with respect to number of training epochs for various embedding training strategies.}
  \label{fig_emb_plot}
\end{figure}

Unsurprisingly, pretrained word embeddings had the greatest impact on performance in terms of both precision and~recall. Early experiments showed that, unlike models from prior work that used static word embeddings~\cite{li2017neural,bekoulis2018adversarial}, our model benefits from trainable word embeddings as shown in Figure~\ref{fig_emb_plot}. Here, trainable word embeddings with \emph{downscaled} gradients refer to reducing the gradient of word embeddings by a factor of 10 at each training step. 

\subsection{Error Analyses}
\label{sec-s-ea}

In this section, we first perform a class based analysis where performance variations for different classes of examples are examined. Then, a more in-depth error analysis is performed for interesting example cases. The class based  analyses entail partitioning examples by length, entity distance, and relation type and are covered in Section~\ref{sec_ea_class}. The more in-depth example based analysis is discussed in Section~\ref{sec_ea_example}.

\subsubsection{Class based analyses}\label{sec_ea_class} 
Long sentences are a natural source of difficulty for relation extraction models given the potential for long-term dependencies. In this section, we perform straightforward  analysis by conducting experiments to assess model performance with respect to increasing sentence length. For this experiment, we train a single model using 80\% of the dataset with 20\% held out for testing. For some sentence length limit $\hat{k}$, we evaluate on a subset of the overall test set that includes only examples with a sentence length that is less than or equal to $\hat{k}$. 

\begin{figure}[ht]
  \begin{minipage}{.48\linewidth}
  \center{\includegraphics[width=\columnwidth]
  {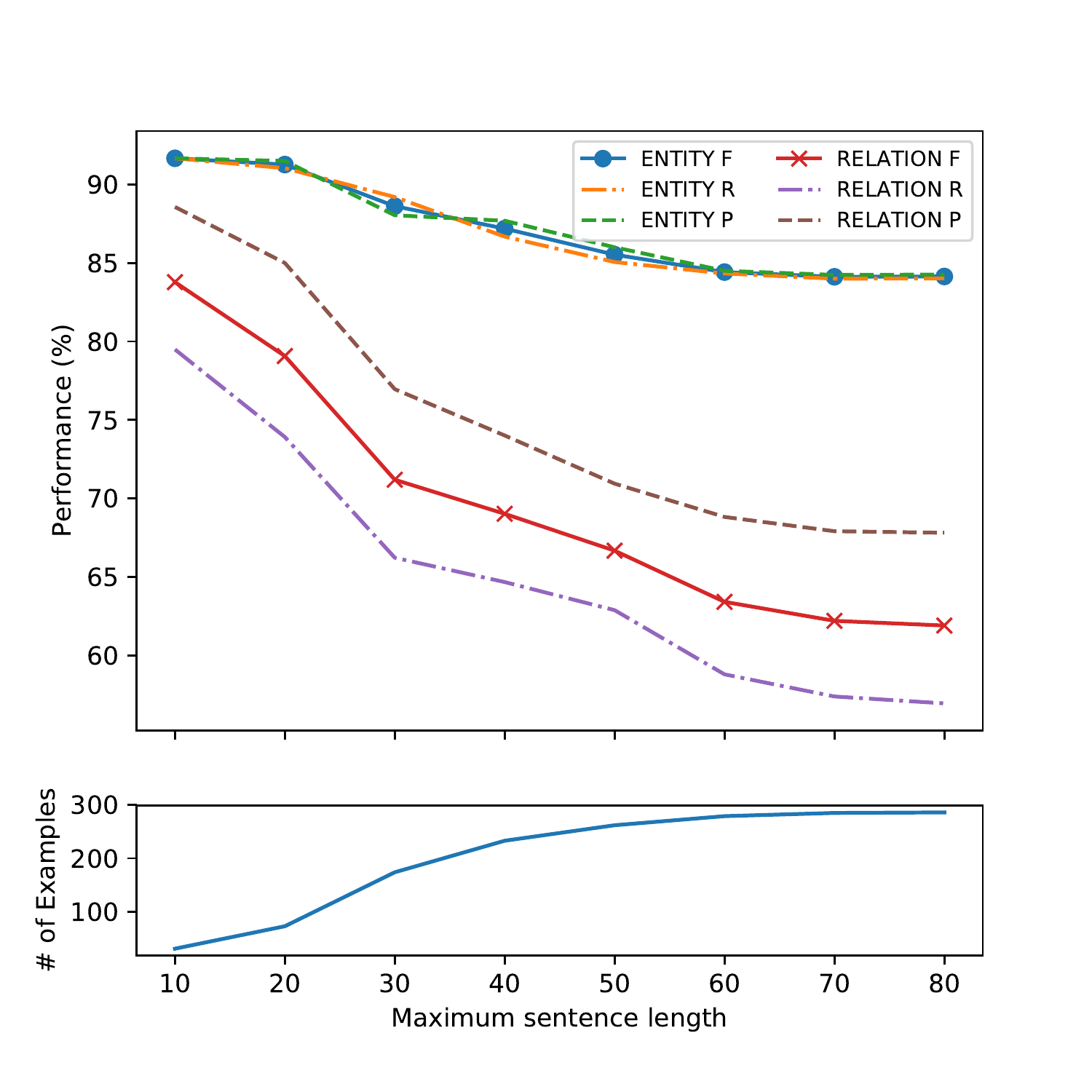}}
  \caption{\textbf{CoNLL04:} Entity and relation extraction performance with respective to change in maximum sentence length.
  }
  \label{fig_conll04_analysis}
  \end{minipage}
  \hfill
  \begin{minipage}{.48\linewidth}
  \center{\includegraphics[width=\columnwidth]
  {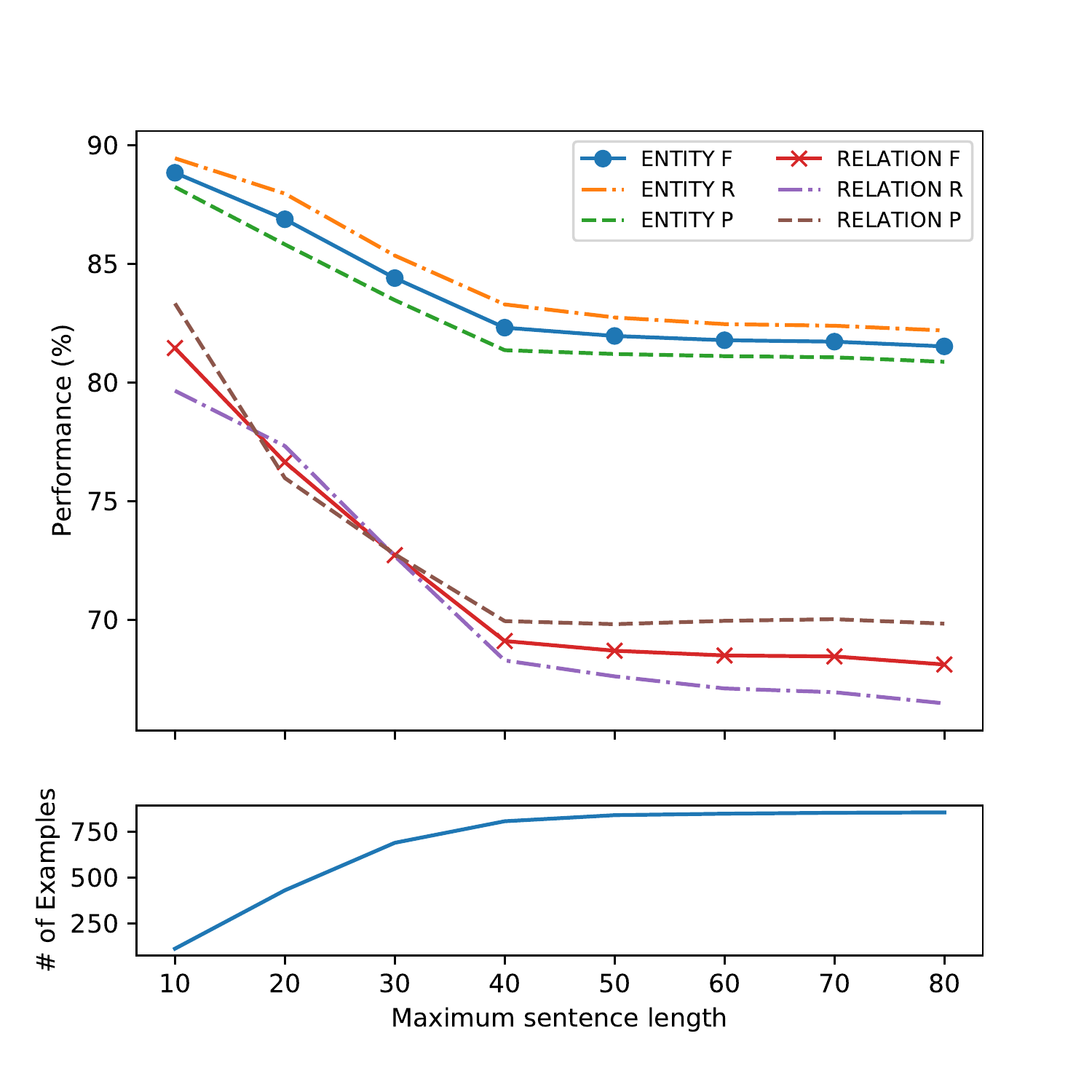}}
  \caption{\textbf{ADE:} Entity and relation extraction performance with respect to change in maximum sentence length.
  }
  \label{fig_ade_analysis}
  \end{minipage}
\end{figure}

Results from these experiments are plotted in Figures~\ref{fig_conll04_analysis} and \ref{fig_ade_analysis}, for the CoNLL04 and ADE datasets respectively, such that $\hat{k}$ is varied along the horizontal $x$-axis. The top graph displays performance, while the bottom graph plots the number of examples with sentence length less than or equal to $\hat{k}$ that are used for evaluation. As shown, performances for both NER and RE tend to decline as longer sentences are added to the evaluation set. Unsurprisingly, relation extraction is more susceptible to long sentences compared to entity recognition. While there is a decline in both relation extraction precision and recall, we note that recall drops at a faster rate with respect to maximum sentence length and this phenomenon is apparent for both datasets. 

\begin{table}[ht]
  \caption{Relation extraction performance partitioned based on ``Entity Distance'', which is defined as the \emph{number of characters} separating the subject and object entities (i.e., absolute character offset).}
  \label{tab_analysis_dist}
  \centering
  \renewcommand{\arraystretch}{1.2}
  \resizebox{\columnwidth}{!}{
  \begin{tabular}{@{\extracolsep{0.2em}}l cccc cccc}
    \toprule
     \, & \multicolumn{4}{c}{CoNLL04 (Relation)} & \multicolumn{4}{c}{ADE (Relation)}\\
    \cline{2-5} \cline{6-9}
     Entity Distance & \# of Examples & \textbf{P} (\%) & \textbf{R} (\%) & \textbf{F} (\%) & \# of Examples & \textbf{P} (\%) & \textbf{R} (\%) & \textbf{F} (\%)\\
     \midrule     
		 0 --- 20 & 207 & 83.7 & 43.80 & 57.51 & 447 & 88.50 & 42.02 & 56.98\\
		20 --- 40 & 51 & 59.09 & 24.07 & 34.21 & 265 & 77.17 & 35.51 & 48.64\\
		40 --- 60 & 43 & 80.00 & 18.60 & 30.19 & 181 & 78.72 & 37.00 & 50.34\\
		60 --- 80 & 22 & 100.00 & 25.93 & 41.18 & 125 & 82.35 & 29.58 & 43.52\\
		80 --- 100 & 13 & 100.00 & 15.38 & 26.67 & 91 & 85.00 & 34.00 & 48.57\\
     \bottomrule
  \end{tabular}}
\end{table}

In addition to length-based analysis, we also conducted experiments to study the variation in relation extraction performance with respect to the distance between subject and object entities as shown in Table~\ref{tab_analysis_dist}. We measure distance by computing the  absolute character offset between the last character of the first occurring entity and first character of the second occurring entity, which is henceforth simply referred to as ``entity distance.'' Our results show that, at least on the CoNLL04 dataset, notable performance differences occur at the boundary cases; i.e., very short range relations (0-20 entity distance) tend to be easier and very long range relations (80-100 entity distance) tend to be harder (mostly due to changes in recall). For the ADE dataset, performance is similar across all partitions of entity distances. This is surprising, as sentence length appears to have a more notable impact on relation extraction performance than entity distance for this particular architecture. 

\begin{table}[ht]
  \caption{Relation extraction performance on the CoNLL04 dataset partitioned based on relation type.}
  \label{tab_analysis_class}
  \centering
  \renewcommand{\arraystretch}{1.05}
  \resizebox{\columnwidth}{!}{
  \begin{tabular}{@{\extracolsep{1em}}l cc ccc}
    \toprule
     \, & \, & \, & \multicolumn{3}{c}{CoNLL04 (Relation)}\\
    \cline{4-6}
     Relation Type & \# of Examples & Avg. Entity Distance & \textbf{P} (\%) & \textbf{R} (\%) & \textbf{F} (\%)\\
     \midrule     
     Kill & 46 & 47 & 81.25 & 82.98 & 82.11\\
     Live\_In & 82 & 37 & 71.76 & 61.00 & 65.95\\
     Located\_In & 58 & 28 & 80.77 & 44.68 & 57.53\\
     Work\_For & 65 & 24 & 60.56 & 56.58 & 58.50\\
     OrgBased\_In & 70 & 29 & 91.38 & 50.48 & 65.03\\
     \bottomrule
  \end{tabular}}
\end{table}

Table~\ref{tab_analysis_class} shows variance in performance when examined by relation type. Here, we see that performance depends heavily on the type of relation being extracted; our model exhibits much higher accuracy on the \emph{Kill} relation at 80\% F1, with \emph{Located\_In} and \emph{Work\_For} being the most difficult with performance below 60\% F1. These results further corroborate our analysis based on Table~\ref{tab_analysis_dist} that entity distance does not correlate with example difficulty given that the \emph{Kill} relation, being the easiest relation to extract, occurs with the highest average entity distance. 

\subsubsection{Example based analysis}\label{sec_ea_example} 
A common source of difficulty that occurs is ambiguity with respect to expression of the  \emph{Live\_In} and \emph{Work\_In} relation types. For example, consider the sentence ``After buying the shawl for \$1,600, \underline{Darryl Breniser} of \underline{Blue Ball}, said the approximately 2-by-5 foot shawl was worth the money.'' The ground truth relation is (Darryl Breniser, \emph{Live\_In}, Blue Ball) which indicates that ``Blue Ball'' is in fact a location. However, it is difficult to assess whether ``Blue Ball'' is a location or company based on the context alone and without broader geographical knowledge (even for humans). Our model predicted (Darryl Breniser, \emph{Work\_For}, Blue Ball) in this case. We observe a similar pattern in the following case: ``\underline{Santa Monica} artist \underline{Tom Van Sant} said Monday after the 23-foot-tall statue was found crushed and broken in pieces.''; here, we see the same phenomenon where our model mistakes (Tom Van Sant, \emph{Live\_In}, Santa Monica) for (Tom Van Sant, \emph{Work\_For}, Santa Monica). Finally, we present the most interesting example of this type of ambiguity in the 
sentence: ```Temperatures didn't get too low, but the wind chill was bad', said \underline{Bingham County} Sheriff's \underline{Lt. Bill Gordon}.'' Here, the ground truth indicates that the only relation to be extracted is (Bill Gordon, \emph{Live\_In}, Bingham County); however, our model extracts (Bill Gordon, \emph{Work\_For}, Bingham County Sheriff), which is also technically a valid relation. Such cases present ambiguities that are also difficult for human annotators; here, imbuing the NER component with external knowledge or learning based on a broader level of context may alleviate these types of errors.

Inconsistencies in the way entities are annotated can also cause issues when it comes to demarcating names that are accompanied with honorifics or titles. For example, some ground truth annotations will include the title, such as ``President Park Chung-hee'' or ``Sen. Bob Dole'', and other cases will leave out the title, such as ``Kennedy'' instead of ``President Kennedy.'' These truth annotations are inconsistent and present a source of difficulty for the model during training and testing. For example, ``\underline{Navy} spokeswoman Lt. \underline{Nettie Johnson} was unable to say immediately whether the aircraft had experienced problems from faulty check and drain valves.'' Here, our model extracted (Lt. Nettie Johnson, \emph{Work\_For}, Navy), while the groundtruth is (Nettie Johnson, \emph{Work\_For}, Navy) --- while both are technically correctly, the extremely precise nature of the evaluation metric causes this prediction to be considered a false positive. 

\begin{figure}[ht]
  \centering{\includegraphics[width=\textwidth]
  {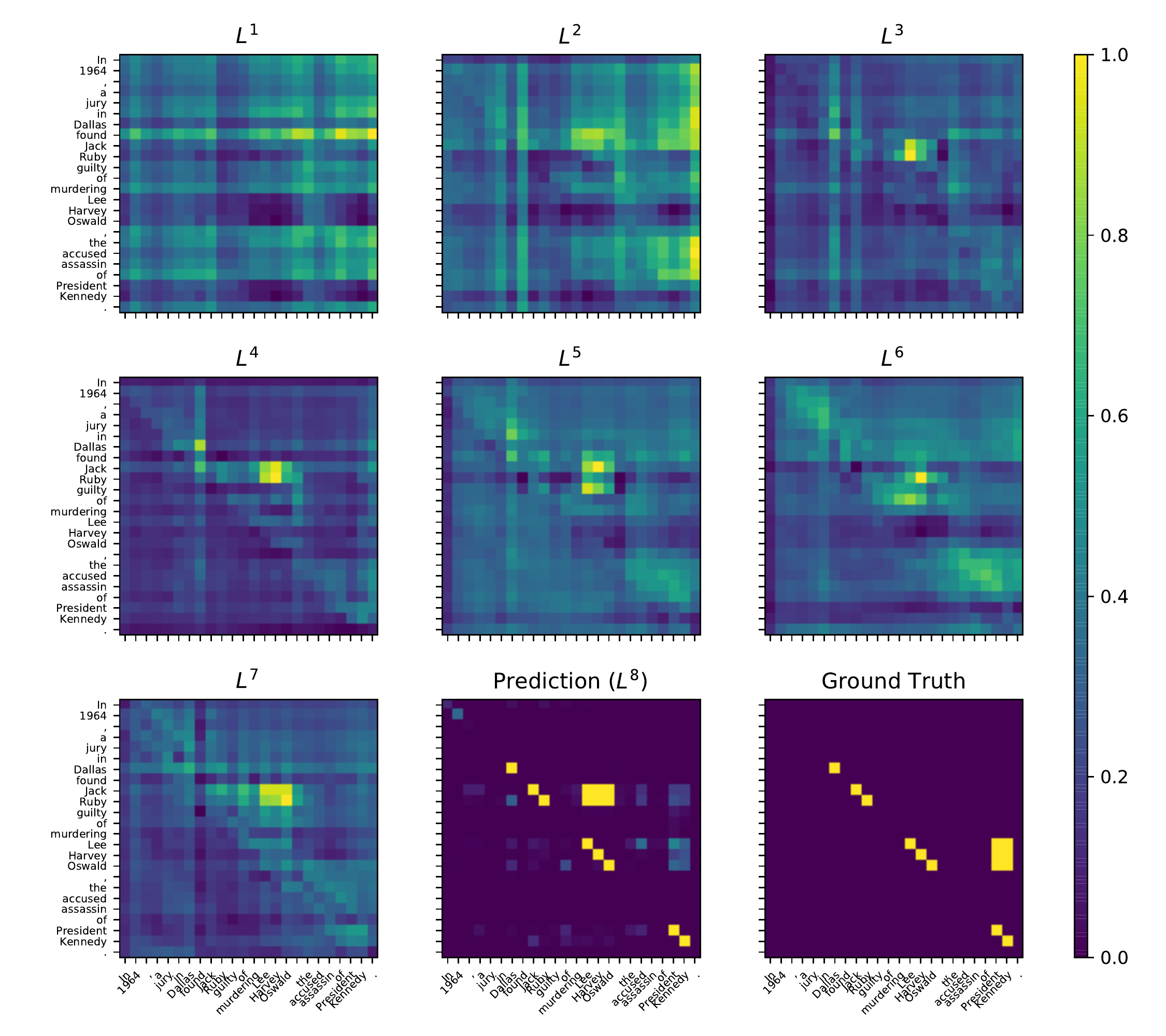}}
  \caption{Visualization of activity of pooling layers at various depths ($L^i$ for $i = 1, \ldots, \lambda$), as tabular heatmaps, for a network with a depth of $\lambda = 8$ given the following input sentence: ``In 1964, a jury in Dallas found Jack Ruby guilty of murdering Lee Harvey Oswald, the accused assassin of President Kennedy.'' Here, we measure activity by sum-pooling the activations along the channel dimension of each hidden representation. For the prediction activity, we simply max-pool probabilities along the relation dimension thus ignoring the exact \emph{type} of entity or relation. 
  }
  \label{fig_kennedy}
\end{figure}

We also see such issues with annotation at the relation extraction stage; for example, consider the sentence ``In 1964, a jury in Dallas found Jack Ruby guilty of murdering \underline{Lee Harvey Oswald}, the accused assassin of \underline{President Kennedy}.'' Figure~\ref{fig_kennedy} shows the internal activity of the network as it attempts to extract entities and relations from this particular example. Here, the ground truth annotation includes (Lee Harvey Oswald, \emph{Kills}, President Kennedy), which our model fails to recognize; we instead obtain the prediction (Jack Ruby, \emph{Kills}, Lee Harvey Oswald) which is a valid relation missed by the ground truth. In fact, it can be argued that the latter relation is a stronger manifested of the ``Kill'' relation based on the linguistic context as evidenced by the trigger phrase ``found [..] guilty of murdering''. We note that our model is able to detect (Lee Harvey Oswald, \emph{Kills}, President Kennedy) as shown in the center-bottom heatmap of Figure~\ref{fig_kennedy}; however, signals were not strong enough to warrant a concrete extraction of the relation.

In the ADE dataset, we mostly observe issues with entity recognition where boundaries of noun phrases are not properly recognized. Modifier phrases are sometimes not predicted as part of the named entity, for example: ``protracted neuromuscular block'' instead extracted as ``neuromuscular block'', and ``generalized mite infestation'' instead extracted as simply ``mite infestation.'' The nature of the data results in especially long named entities that are often entire noun or verb phrases which can be difficult to delimit. For example, consider the following case: ``DISCUSSION: Central nervous system (CNS) toxicity has been described with ifosfamide, with most cases reported in the pediatric population.'' Here, instead of extracting (Central nervous system (CNS) toxicity, ifosfamide) as the relation pair, our model predicts (Central nervous system, ifosfamide) and (CNS, ifosfamide). Essentially, long entity phrases are often  not recognized in their entirety, and broken down into segments where each segment is independently involved in a relation. In this particular case, this error in prediction lead to one false negative and two false positives. This phenomenon occurs frequently with coordinated noun phrases which present a nontrivial challenge. For example, ``\underline{Growth and adrenal suppression} in asthmatic children treated with high-dose fluticasone propionate.'' is annotated with ``Growth and adrenal suppression'' as a singular entity, while our model falsely recognizes it as two entities ``Growth'' and ``adrenal suppression.'' We see similar outcomes for the sentence: ``\underline{Generalized maculopapular and papular purpuric eruptions} are perhaps the most common thionamide-induced reactions.'' Such cases occur frequently which we suspect are a major source of hampered precision given the increased number of false positives for each predictive mistake.

\section{Conclusion}
In this study, we introduced a novel neural architecture that combines the ideas of metric learning and convolutional neural networks to tackle the highly challenging problem of end-to-end relation extraction. Our method is able to simultaneously and efficiently recognize entity boundaries, the type of each entity, and the relationships among them. It achieves this by learning intermediate table representations by pooling local metric, dependency, and position information via repeated application of the 2D convolution. For end-to-end relation extraction, this approach improves over the state-of-the-art across two datasets from different domains with statistically significant results based on examining average performance of repeated runs. Moreover, the proposed architecture operates at substantially reduced training and testing times with testing times that are seven to ten times faster, the latter important for many user-end applications. We also perform extensive error analysis and show that our network can be visually analyzed by observing the hidden pooling activity leading to preliminary or intermediate decisions. Currently, the architecture is designed for extracting relations involving two entities and occur within sentence bounds; handling $n$-ary relations and exploring  document-level extraction involving cross-sentence relations will be the focus of future work. 

\starttwocolumn
\bibliography{scoonerdb}
\end{document}